# MIXED FORMAL LEARNING
A Path to Transparent Machine Learning

Sandra Carrico
Vice President & Chief Data Scientist
Glynt.AI
Mountain View, CA. USA

*Abstract*—**This paper presents Mixed Formal Learning, a new architecture that learns models based on formal mathematical representations of the domain of interest and exposes latent variables. The second element in the architecture learns a particular skill, typically by using traditional prediction or classification mechanisms. Our key findings include that this architecture: (1) Facilitates transparency by exposing key latent variables based on a learned mathematical model; (2) Enables Low Shot and Zero Shot training of machine learning without sacrificing accuracy or recall.** *(Abstract)*

*Keywords-component; explainability, transparency, mixed formal learning, machine learning, low shot, zero shot*

## I. INTRODUCTION

Today, machine learning models that utilize only neural networks lack transparency. Neural networks, or other machine learning methods combined with formal mathematical models addresses the lack of transparency, while retaining the option of utilizing otherwise opaque models. We call this architecture Mixed Formal Learning. It exposes latent variables based on a formal mathematical model of the domain with semantic meaning. Latent variables enable the development of families of learned-models relevant to the domain. The use of formal mathematical representations to facilitate learning a model to expose latent variables, in part, addresses transparency.

This method also produces the effect of Low and Zero Shot learning. This paper defines Low Shot learning as supervised training requiring 10 or fewer examples. Zero Shot learning is defined as requiring no training examples for tasks that are traditionally supervised. Our GLYNT application requires fewer than 10 labeled examples in many cases to extract numerical and phrase data from semi-structured documents with F1 scores of 95% or better.

## I. RELATED WORK

Though our work with the GLYNT system has gone a step further, recent work at other companies has parallels to our own. Wengong Jin, et al. [1] used an autoencoder to find latent variables for their second model to automate the design of molecules based on specific chemical properties. The autoencoder, while not apparently based on a formal model, enables 100% validity of identified molecules, an effect consistent with our performance expectations of a well-designed formal model. R Devon Hjelm, et al. [2] explore mutual information, a concept that is related to how a formal model helps to guide the second model. Their paper also uses adversarial techniques to computationally learn a model of a latent space of variables. This differs from our method of defining the formal model a-priori. Jacob Devlin, et al. [3] create a language representation model to enable a number of fine-tuning models for different applications. The fine-tuning models correspond to the non-formal aspects of our architecture.

A paper by Supasorn Suwajanakorn et al. [4] of Google follows our Mixed Formal Learning architecture. The paper identifies key points, spatial locations that stand out on an image and remain regardless of distortion. The authors describe training two neural networks. The first network uses the orthogonal Procrustes problem as a learning method to create a 3D representation from two sequential 2D images, such as photographs. In other words, imagine two successive still-frames of a Ferrari in motion. Insert these two flat images into Google's system and it can render a representation of a 3D image of the Ferrari.

The 3D representation enables the next model to more easily find keypoints by exploiting the helpful latent variables. The second model's more typical neural net finds keypoints as accurately, or better, by using the exposed latent variables, rather than solutions utilizing large sets of labeled training data. Since the method required no labels on the training data, the result is image keypoint identification with Zero Shot Learning.

At the conceptual level, GLYNT and Google used the same approach: an architecture of multiple models, one to model the domain and one to make predictions from the first model's outputs. The synergy between the two models often results in Zero or Low Shot training requirements.

## II. MIXED FORMAL LEARNING IMPLEMENTATION

The use of the formal representation to generate latent variables exposes key features of the problem domain that other models leverage. These exposed variables provide a level of semantic understanding of the family of models and facilitates transparency. The transparency helps humans understand the model and also exposes that understanding to subsequent models. In addition, the implementation benefits include faster training on less data, with potentially excellent accuracy.

In the GLYNT application of Mixed Formal Learning, the architecture learns a domain-specific mathematical model for semi-structured documents and then combines exposed latent variables with typical machine learning methods based on observed data. The result is an AI solution that requires small training sets while often achieving F1 scores of 95% or better. The GLYNT application of Mixed Formal Learning requires fewer than 10 examples for training to extract specific fields.

While Mixed Formal Learning can support any number of models, the GLYNT implementation uses two primary models arranged in sequence. The first model learns latent variables based on a tightly constructed mathematical representation, tailored to the domain. Getting this representation right is key to enabling Low Shot Learning. The next model in the sequence combines the latent variables and the incoming labeled data and learns to find a phrase or numerical field and properly assign a label to it.

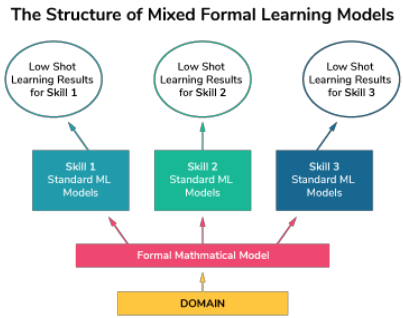

Figuratively speaking, to learn from such little data, the formal math-based model shines a glint of light on the area of the answer, allowing the second model to more quickly learn its skill. It was this glint effect that is the origin of the name GLYNT.

The formal model enables a set of capabilities shared by diverse domains. Each domain can leverage the formalism to create AI applications. GLYNT, for example, uses a domain-specific model to extract unstructured data, but the formal mathematical model can also be used in supply chain optimization and genomics data sets.

The second element in the architecture learns a particular skill, typically by using traditional prediction or classification mechanisms. A fine-tuning approach, as is often used with embeddings. It would qualify as a Mixed Formal Learning architecture if the embeddings were produced based on a transparent mathematical formalism rather than an empirically successful happenstance. The initial GLYNT application identifies fields according to customer-specified names.

It is notable that the formal model enables follow-on applications. Zero Shot applications should be expected for some cases. In the case of GLYNT, an application of interest would identify new fields that alert users to field-level changes in their set of documents. We envision a Zero Shot implementation in this case. Another useful follow-on application would categorize the documents according to the publisher.

### A. Sources for formal model inspiration

Operations research and avionics historically used formal models for prediction. Their literature provides many novel applications. Typically, those methods are limited by the accuracy of the mathematics. Mixed Formal Learning improves on bare mathematical models, which often imperfectly model the domain, with a machine learning based model. The result benefits from the transparency of a formal model with the power of modern machine learning techniques.

The architecture also benefits from the ability to base a family of applications related to the domain upon the learned formal model. Businesses typically need to predict or classify many aspects of their data set. Once the formal domain model and its generated latent variables are learned, additional models can learn quickly in Low Shot mode with few labels, or even with no labels, for models that otherwise would require supervised learning techniques

### B. Potential Formal Models

Thus far we have cited two examples of real-world solutions using the Mixed Formal Learning architecture. How can others confidently apply this method to solve their own problems? What are some potential formal models? We examined a few examples below as exemplars. The practitioner should deeply understand the domain and the business needs to find ideal formal models.

#### 1) Self driving Cars

Understanding the movement of many objects in the real world is useful for applications such as self-driving cars. While Newtonian physics provides a straightforward model that could expose such latent variables to a model, it may prove impractical to calculate the trajectory for many moving objects in space. In that case learning a model based in part on Newton physics to predict the location of many objects may perform better on small embedded chips as might be found in cars.

The secondary models in that case may predict the best evasion maneuver if the formal model predicts a collision. In that case the secondary model would need to consider several angles of steering or breaking.

#### 2) Supply Chain Optimization: Picking

When picking and packing for a warehouse, the locale of the items influences the efficiency and capacity of the operation. Fluid flow models used to optimize a velocity based stowage policy could be used to learn a model to optimally place objects within the warehouse [5]. That base-model could inform models when determining optimal warehouse size, placement, supply and resupply.

#### 3) Improved Recycling Center Design

A multi-stage material separation process such as might be used in cracking can be repurposed to model operations within a recycling plant. This formal model enables learning a useful set of latent variables that closely match the domain semantic[5]. With those latent variables additional models could predict output of the recycling center, expected revenues, and expected mixes of outputs.

### III. EXPERIMENTS

To demonstrate the performance of the GLYNT mixed formal learning implementation, we ran a series of tests to extract data from semi-structured documents. The first observation was that the formal model delivered such high-quality latent variables to the subsequent model that the combination achieved better than 95% F1 score with 7 or fewer training examples. This capability is often referred to as Low Shot learning.

The graph below shows results from the extraction of unstructured data from four document sets, each with 50 documents. The documents are from the energy industry (2 different utility bills) and the healthcare industry (insurance cards and clinic visit summaries). 10 documents in each set were reserved for training and 40 documents in each set form a holdout set. The graph shows how well the data extraction performs as the size of the training document set increases (For more details see our related paper: Low Shot Learning in Action).

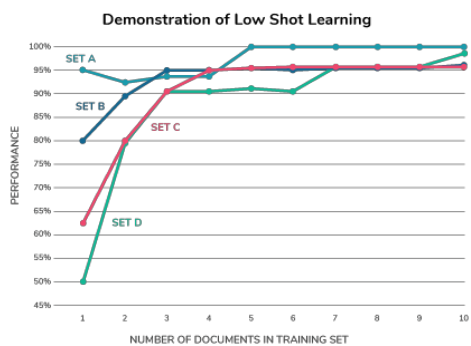

The F1 score is displayed on the graph for each document set. The key result is F1 score performance at 95% and above, obtained with fewer than 10 documents in the training set. As the graph indicates, the gain in performance levels off at about 7 documents. Most of the improvement in performance derives from improvements in recall, as accuracy remains fairly uniform. In previous work, we have documented the high performance of the GLYNT data extraction system. The results here show that with the addition of Low Shot Learning, GLYNT is able to achieve this level of performance on very little training data. This is in sharp contrast to other machine learning systems that require training data sets of 15,000 to 300,000 documents and months of work for similar tasks. GLYNT produces results in under an hour

## IV. CONCLUSIONS

Mixed Formal Learning liberates unstructured data more transparently with significantly less training data. As well, it eliminates the advantage of large data sets in the AI ecosystem and creating a new landscape where advanced AI products and solutions are available to researchers and data scientists at large, rather than only the lucky few sitting on huge data assets.

We expect to see three extensions of this work. First, the formal mathematical models of Mixed Formal Learning is tightly tailored to the domain at hand, but with mathematical insight, the same model can be used in seemingly disparate domains. As mentioned, the GLYNT formal model might well capture aspects of the domains of genomic data analysis and supply chain optimization. Second, other researchers are edging towards the same architecture, and enabling Low Shot Learning in structurally different applications. Third, the second model of Mixed Formal Learning provides automated skills. In the GLYNT application the skill is extraction of unstructured data. Additional skills can be used with GLYNT's formal model, such as new field identification or document classification. It is expected that the new skills will have Zero or Low Shot functionality and require very little training data if any.

ACKNOWLEDGMENT

Thank you to Martha Amram for her work editing this work, and setting up and executing the experiments.